\documentclass[journal]{IEEEtran}

\usepackage{amssymb, amsmath, amsthm, graphicx, enumerate}
\usepackage{setspace}
\usepackage{algorithm}
\usepackage{algorithmicx}
\usepackage{array}
\usepackage{float}
\usepackage{amstext}

\usepackage{graphicx}
\usepackage{booktabs}
\usepackage{tabularx}
\usepackage{multirow}
\usepackage{cite}
\usepackage{stfloats}
\usepackage{cite}
\usepackage{graphicx}
\usepackage{arydshln} 
\usepackage{verbatim}
\usepackage{bbm}
\usepackage{stfloats}
\usepackage[justification=centering]{caption}
\usepackage{bm}
\usepackage[noend]{algpseudocode}
\usepackage{caption,subcaption}
\usepackage{epstopdf}
\usepackage{makecell}
\usepackage{url}
\usepackage{lineno}
\begin{document}

\title{Bayesian Layer Graph Convolutional Network for Hyperspectral Image 
Classification }

\author{Mingyang Zhang, \IEEEmembership{Member, IEEE},  Ziqi Di,  Maoguo Gong, 
\IEEEmembership{Senior Member, IEEE}, \\ Yue Wu, \IEEEmembership{Member, IEEE}, 
Hao Li, \IEEEmembership{Member, IEEE}, Xiangming Jiang, \IEEEmembership{Member, 
IEEE}

\thanks{M. Zhang, Z. Di, M. Gong, H. Li and X. Jiang are with the School of 
Electronic Engineering, and the Key Laboratory of Intelligent Perception and 
Image Understanding of Ministry of Education, Xidian University, No. 2 South 
TaiBai Road, Xi’an 710071, China (e-mail: omegazhangmy@gmail.com; 
gong@ieee.org).

Y. Wu is with the School of Computer Science and Technology, Xidian University, 
Xi’an 710071, China.}

}


\maketitle

\begin{abstract}
In recent years, research on hyperspectral image (HSI) classification has 
continuous progress on introducing deep network models, and recently the graph 
convolutional network (GCN) based models have shown impressive performance. 
However, these deep learning frameworks are based on point estimation suffering 
from low generalization and inability to quantify the classification results 
uncertainty. On the other hand, simply applying the distribution estimation 
based Bayesian Neural Network (BNN) to classify the HSI is unable to achieve 
high classification efficiency due to the large amount of parameters.
In this paper, we design a Bayesian layer as an insertion layer into point 
estimation based neural networks, and propose a Bayesian layer graph 
convolutional network (BLGCN) model by combining graph convolution 
operations, which can effectively extract graph information and estimate the 
uncertainty of classification results.
Moreover, a Generative Adversarial Network (GAN) is built to solve the sample 
imbalance problem of HSI dataset. 
Finally, we design a dynamic control training strategy based on the confidence 
interval of the classification results, which will terminate the training early 
when the confidence interval reaches the presented threshold.
The experimental results show that our model achieves a balance between high 
classification accuracy and strong generalization. In addition, it can 
quantify the uncertainty of outputs.

\end{abstract}


\IEEEpeerreviewmaketitle

\section{Introduction}
\label{Introduction}
\IEEEPARstart{W}{ith} the rapid development of remote sensing sensors, 
hyperspectral images have drawn more and more attentions in academic research 
and various applications \cite{he2017recent, zhu2017deep}, such as mineral 
detection, soil testing, medical image, and urban planning. Compared with 
conventional optical remote sensing image, hyperspectral images contain 
abundant spectral information, which are consists of hundreds of contiguous 
spectral channels \cite{zhang2016deep, chang2007hyperspectral}. Due to this 
character, besides spatial context information, hyperspectral images possess 
unique spectral signatures for identifying materials and ground truth. For the 
last few years, spatial-spectral classification for hyperspectral images has 
become a hot research issue in remote sensing society.  

In early researches, hyperspectral image classification focuses on designing 
various feature extraction methods manually. Feature extraction is designed to 
mapping the original data into a new feature space where the spatial and 
spectral features are more distinguished and better for classification. For 
spectral feature extraction, there are two main methods: linear transformation 
based methods and nonlinear transformation based methods. Principal component 
analysis (PCA) \cite{shlens2014tutorial} and independent component analysis 
(ICA) \cite{dalla2011classification} are two representative methods for linear 
transformation based methods. They use matrix factorization to obtain a new 
feature space. For nonlinear transformation based methods, locally linear 
embedding (LLE) \cite{Roweis2000Nonlinear} and locality preserving projections 
(LPP) \cite{he2004locality} based methods are two conventional feature 
extraction methods in pattern recognition, which are based on manifold learning 
for seeking new feature space. However, researchers found that only extracting 
spectral features has limited improvement in classification accuracy. Since 
spatially adjacent ground truth is more relevant \cite{tobler1970computer}, 
spatial context features are also considered for feature extraction combined 
with spectral features. Random field based methods are classic spatial feature 
extraction methods. Thereinto, Markov random field (MRF) and conditional random 
field (CRF) are two representative ones. MRF can build models according to the 
relevance of adjacent pixels, which can be used in multiple ground truth scenes 
and multiple  distribution models. Many works have been proposed based on MRF. 
For example, an adaptive MRF was proposed for hyperspectral feature extraction 
\cite{zhang2011adaptive}. Further, Sun proposed a spatially adaptive Markov 
random field (MRF) prior in the hidden field \cite{sun2014supervised}, which 
makes the spatial term more smooth. CRF is a probabilistic model for labels and 
structural data, which shows potentials for spatial feature extraction. Various 
CRF-based methods have been proposed for spatial feature extraction, such as 
CRF combined with sparse representation \cite{zhong2010modeling}, CRF combined 
with ensemble learning \cite{li2015hyperspectral}, and CRF combined with 
sub-pixel techniques \cite{zhao2015sub}. Besides, morphological feature based 
methods are also effective ways for spatial feature extraction, among which, 
extended morphological attribute profiles (EMAP) is a representative one 
\cite{plaza2004new}. Based on EMAP, some solid works have been proposed, such 
as multiple feature learning \cite{li2015multiple} and histogram-based 
attribute profiles methods \cite{demir2015histogram}.

The conventional manually designed feature extraction methods consist of 
monolayer structure generally, which can be considered as shallow models. 
However, hyperspectral images have shown the characteristics of nonlinear, high 
dimensions and redundancy. Due to the capacity limitation of shallow models, it 
is always not ideal for shallow models when processing this type of data. To 
address this issue, the deep learning models have been introduced to 
hyperspectral image classification and some inspiring works have been proposed 
to implement spatial-spectral classification in recent years. Chen first 
proposed to introduce a deep learning model to hyperspectral classification 
\cite{chen2014deep}, which using a stacked auto-encoder structure to extract 
high level features. Based on this work, a deep belief network was proposed to 
extract spatial-spectral features for hyperspectral images 
\cite{chen2015spectral}. After this, various convolutional neural network (CNN) 
based methods started to make their marks in hyperspectral classification, due 
to their great potentials for extracting spatial features. A 2D and 3D CNN 
framework were proposed to implement classification for hyperspectral images 
with a supervised style \cite{chen2016deep}, which further improves the 
classification performance. Lee proposed a contextual CNN with a deeper 
structure to extract more subtle spatial features for classification 
\cite{lee2017going}. A diverse region-based CNN was proposed to obtain 
promising features with semantic context-aware representation 
\cite{zhang2018diverse}. Recently, a CNN with multiscale convolution and 
diversified metric framework was proposed, which combines multiscale learning 
and determinantal point process priors to improve the diversity of features for 
classification \cite{gong2019cnn}. However, CNN based methods bring another 
challenge for hyperspectral images classification. The absence of labeled 
samples can not sufficiently support the training of CNN based models. To 
address this issue, data augmentation and unsupervised pre-training are 
introduced to CNN based methods. For data augmentation, 
Li proposed a pixel-pair feature extraction strategy to enlarge the labeled 
samples \cite{li2017hyperspectral}. With the enlarged labeled samples, the 
training of deep CNN model can be improved effectively. For unsupervised 
pre-training strategy, Romero proposed a greedy layerwise training framework to 
implement unsupervised pre-training \cite{romero2016unsupervised}. Further, Mou 
proposed an encoder-decoder structure to train a residual CNN block in an 
unsupervised style \cite{mou2017unsupervised}. Moreover, a generative model 
based on Wasserstein generative adversarial network was proposed 
\cite{zhang2018unsupervised}, which can pre-train a CNN subnetwork without 
labeled samples. 

Recently, besides Euclidean-space based deep learning methods, the graph-space 
based deep learning methods have attracted more attention for the research of 
hyperspectral image classification, which maps the original data from Euclidean 
space to graph space. Due to the concepts of nodes and edges, compared 
with traditional Euclidean space, the graph space can better explore and 
exploit the relationship among the whole pixels of a hyperspectral image 
\cite{sellars2020superpixel}. Based on this feature, some graph neural network 
methods have been proposed for the hyperspectral image classification task. Wan 
transformed hyperspectral data to a graph style using image segmentation 
techniques, and proposed a multiscale dynamic graph convolutional network (GCN) 
to implement node classification \cite{wan2019multiscale}. Based on 
\cite{wan2019multiscale}, a context-aware 
dynamic graph convolutional network was proposed, which can flexibly explore 
the relations among graph-style data \cite{wan2020hyperspectral}. Meanwhile, 
due to the ability of handling the whole graph i.e. the whole data, some 
semi-supervised graph-based methods such as GCNs have been proposed, which can 
utilize the labeled and unlabeled samples simultaneously to reduce the 
requirement of labeled training samples \cite{qin2018spectral, 
sellars2020superpixel, mou2020nonlocal}. Moreover, some effective variants of 
GCNs have been further designed. In \cite{he2021dual} and \cite{wan2021dual}, 
GCNs have been designed in dual channel style to explore multiscale spatial 
information and label distribution respectively. For better update the graph 
structure, in \cite{ding2021adaptive} and \cite{bai2021hyperspectral}, an 
adaptive sampling strategy and an attention mechanism are combined with the 
GCNs-based methods respectively. Liu combined the advantages of convolution 
neural networks with GCNs to better utilize pixel and superpixel level features 
\cite{liu2020cnn}. From the view of training efficiency, Hong proposed a 
minibatch GCN which can reduce computational cost in large scale remote sensing 
data analysis \cite{hong2020graph}. 

However, the aforementioned deep learning-based methods, including CNN and GCN, 
use point estimation for data features, which makes the network easy to get 
overfitted with poor generalization ability during the learning process. At the 
same time, since each forward propagation is performed with a single sampling 
of the weight space, the convergence speed and learning efficiency are greatly 
limited, and it is incapable for the network to estimate the credibility of its 
output. 
In addition, in the face of the imbalanced sample distribution of 
hyperspectral remote sensing datasets, it is difficult for deep learning 
methods to maintain high classification accuracy on minority classes, 
which increases the risks of miss alarming corresponding to high-value samples.

To address the issues discussed above, we propose a Bayesian layer graph 
convolutional network (BLGCN), in which we design the Bayesian layer. 
Different from the traditional neural network, Bayesian strategy represents 
weight matrices in distribution form, which transforms the training process of 
networks from point estimation to distribution estimation. The weight matrices 
we get from sampling are combined with input features to obtain the network 
output, i.e., the classification results. 

We apply the variational inference method to calculate the posterior 
distribution of the weight matrices and classification loss. Consequently, 
based on the given prior distribution and multiple sampling in each forward 
propagation, the overfitting problem is effectively solved, and the 
generalization ability of the neural network is significantly improved.

While adapting the Bayesian neural network, we give the definition of Bayesian 
layer and design its structure. The participation of Bayesian layer enables  
researchers to fine-tune the network backbone to satisfy the specific task. 
Compared with the traditional Bayesian neural network, our method can reach 
the balance of high classification accuracy and low time cost. 

Besides, in order to tackle the task of minority class recognition, 
generative adversarial methods \cite{goodfellow2014generative} are introduced  
to perform feature learning and data generation for minority class. By 
expanding the training set with generation data, the recognition accuracy of 
minority class is effectively increased.

At the same time, for the purpose of using the uncertainty information of 
classification results to guide the training process, we design a dynamic 
control strategy with Bayesian method. By setting the threshold on the 
validation set classification accuracy, it is stipulated that the training 
process will be terminated when the upper limit of the confidence interval 
reaches the threshold. This strategy shortens the time cost of training process 
and increases the training efficiency.

To sum up, our main contributions in this paper are listed as follows:

(1) We propose the BLGCN framework, which can reach the balance of high 
classification accuracy and low time cost.

(2) The Bayesian method carried by Bayesian layer enables the network to 
quantify the uncertainty of the classification results.

(3) The generative adversarial method we import solve the minority class 
recognition problem efficiently.

(4) A dynamic control strategy based on Bayesian method is involved in the 
training process which further reduces the time cost of model learning.

The remainder of our paper is organized as follows. Background and motivations 
of the methods we used are introduced in Section II. In Section III we 
formulate the theoretical basis of BLGCN and relevant image processing methods, 
including data-augmentation with generative adversarial methods and our 
training strategy. We conduct the experiments in Section IV. Finally, this 
article is concluded in Section V.

\section{Background and Motivation}
\label{Background and Motivation}
\subsection{Graph Convolutional Network}

Since the graph convolutional networks presented by Kipf\&Welling in 2017 
\cite{kipf2016semi}, it has been introduced into a wide range of applications, 
including recommendation system, graph embedding and node classification. As an 
important part of remote sensing images analysis, the application of GCN 
on hyperspectral image classification can also be seen in several works 
\cite{hong2021revisiting}, \cite{yang2021spatial}, \cite{chen2021automatic}. To 
begin with, the GCN method assumes that the given settings can be simplified 
into a graph $G_{obs}=(V,\varepsilon)$, where $V$ represents the set of $N$ 
nodes and $\varepsilon$ is the set of their edges. Each node $i$ has a set of 
feature vectors $x_i\in \mathbbm{R}^{d\times 1}$ in $d$ dimensions, and the 
label set related to a subset of nodes $L\subset V $ can be expressed as $y_i$. 
In the image classification task, the value of label set identifies categories.

The core layer of a GCN can be expressed as follows:
\begin{equation}
\begin{aligned}
\bold{H}^{(1)}&=\sigma(\hat{\bold{A}}_G\bold{X}\bold{W}^{(0)}), \\
\bold{H}^{(l+1)}&=\sigma(\hat{\bold{A}}_G\bold{H}^{(l)}\bold{W}^{(l)}).
\label{GCN}
\end{aligned}
\end{equation}

Here $\bold{W}^{(l)}$ denotes the weights of the $l$th layer 
in a GCN. $\bold{H}^{(l)}$ represents the output features from $l$th layer, 
and the function $\sigma$ is a nonlinear activation function. The normalized 
adjacency matrix $\hat{\bold{A}}_G$ is derived from a given graph and 
determines the mixing situation of the output features across the graph at each 
convolutional layer. 

For an $L$-layers GCN, the final output is $Z= \bold{H}^{(L)}$. The model 
learns the weights of network through the backpropagation with the target of 
minimizing the error metric function between the given labels $Y_L$ and the 
network prediction outputs $Z_L$.

\subsection{Bayesian Method}
With the development of convolutional neural networks and Transformers in 
recent years, mainstream deep learning frameworks have become more and more 
complex, and the network depth can reach hundreds or even thousands of layers. 
Although these large-scale neural networks are capable of information 
perception and feature extraction, there still remains hidden dangers of easy 
overfitting and lack of the ability to estimate credibility of the model 
output. 
Hence, Bayesian deep learning method \cite{goan2020bayesian} 
has been widely used in research and gets a significant effect since it can 
capture the cognitive uncertainty inherent in data-driven models as well as 
maintaining a high accuracy\cite{shridhar2019comprehensive}.

For the given dataset $\bm{X}$ and label set $\bm{Y}$, when predicting the 
probability distribution of the test data pair $x^*$ and $y^*$, according to 
the marginal probability calculation, we have
\begin{equation}
p(y^*\mid x^*,\bm{Y},\bm{X})=\int p(y^*\mid x^*,\omega)p(\omega\mid\bm{Y},\bm{X}) \label{marginal 
probability}
\end{equation}
where $\omega$ is the model parameter, and the problem is transformed into 
finding the maximum posterior distribution of the parameters on the training 
dataset $\bm{X}$,$\bm{Y}$. According to the 
Bayesian formula, we have
\begin{equation}
p(\omega\mid\bm{Y},\bm{X})= \frac{p(\bm{Y}\mid \bm{X},\omega)p(\omega)}{p(\bm{Y}\mid\bm{X})} 
\label{Bayesian formula}.
\end{equation}

The integral in Eq. \ref{marginal probability} is intractable for most 
situations, and there are various integral methods proposed to infer the 
$p(\omega\mid\bm{Y},\bm{X})$ including variational inference 
\cite{graves2011practical} and Markov Chain Monte Carlo (MCMC) method 
\cite{goan2020bayesian}.

The general Bayesian deep learning is defined slightly differently in various 
articles, but it usually refers to the Bayesian neural network. Based on the 
framework of the traditional convolutional neural network, Bayesian neural 
network inherits the method of Bayesian deep learning. Assuming the weight 
$\omega$ subjects to a specific distribution, the network 
optimizing goal is to maximizing the posterior distribution 
$p(\omega\mid\bm{Y},\bm{X})$. The backpropagation process remains the same with 
traditional neural network. 

By structurally combining probabilistic model and neural network, Bayesian 
neural network also keeps their advantages. While retaining the ability of 
neural network perception and feature extraction, Bayesian method applies 
distribution estimation instead of point estimation by representing the weight 
space with probability distribution. It provides theoretical premise for 
quantifying the confidence level of the learning results. By measuring the 
uncertainty of learning results, we can design evaluation models, which 
dynamically evaluate the training progress and determine whether to proceed. At 
the same time, the training method based on distribution estimation speeds up 
the learning convergence process and ensures the robustness of the training 
results by sampling the weight distribution multiple times in a forward 
propagation process.

\subsection{Generative Adversarial Network}
Nowadays, GAN \cite{goodfellow2014generative} is becoming a popular model with 
strong generative capabilities and high efficiency. Generally, it consists of a 
discriminative network $\bm{D}$ and a generative network $\bm{G}$. The 
generative network $\bm{G}$ needs to generate features that match the training 
features $\bm{x}$ as much as possible based on the given random features 
$\bm{z}$, which subjects to a specific distribution (e.g., uniform 
distribution or the Gaussian distribution). On the contrary, the aim of the 
discriminative network $\bm{D}$ is to distinguish whether the sample data is 
generated features or training features $\bm{x}$. The training processes of the 
two networks are arranged to perform alternately, which can be considered as a 
two-player mini-max game and has a value function as follows:
\begin{equation}
\begin{aligned}
\min_{\bm{G}}\max_{\bm{D}}V(\bm{D},\bm{G})&=\mathbbm{E}_{\bm{x}\sim 
p_{data}(\bm{x})}[\log 
\bm{D}(\bm{x})]\\
&+\mathbbm{E}_{\bm{z}\sim p_{\bm{z}}(\bm{z})}[\log 
(1-\bm{D}(\bm{G}(\bm{z})))].
\end{aligned}
\end{equation}

It is assumed that the training features $\bm{x}$ subject to the distribution 
of $p_{data} (\bm{x})$, and the random features $\bm{z}$ subject to an 
arbitrary distribution $p_{\bm{z}} (\bm{z})$. As training processes, the 
discriminator network and the generative network tend to converge and reach a 
point where the features generated are so real that the discriminator is unable 
to judge whether they are real features.

When processing hyperspectral remote sensing image data, researchers are often 
troubled by the problem of sample imbalance. In the Indian Pines dataset, the 
sample size of the smallest class is only equals to eight thousandths of the 
sample size of the largest class. The resulting model will easily have poor 
classification performance on the minority class. If we simply replicate the 
minority class data, we can achieve high recognition accuracy on the training 
dataset, but the model we got will lack generalization ability when 
dealing with unknown datasets. 
Therefore, we need to design the model from the data generation point of view. 
The traditional generative adversarial network has strong generative ability 
and adaptability, but when dealing with the pixel data with long-sequence 
features in hyperspectral images, the feature extraction ability will be 
greatly reduced, and become difficult to generate new features. 

Over the past years, GAN methods including 1D-GAN \cite{zhu2018generative}, 
3D-GAN \cite{zhu2018generative} and other networks \cite{yin2019hyperspectral}, 
\cite{wang2019semi}, \cite{zhan2017semisupervised} have demonstrated their 
superior ability in unsupervised or semi-supervised learning, pioneers have 
started to apply them in dealing with hyperspectral remote sensing problems.
However, we find that these models have limited feature expression capabilities 
and overfitting problem. The feature extraction method also loses part of the 
original information.

Therefore, we make some improvements to the generative adversarial network to 
solve the problem. By means of feature vector interleaved combination and unit 
matrix feature extraction, the newly designed generative adversarial network 
achieves better feature expression ability and strong generalization at the 
same time.

\section{Methodology}

This section elaborates our proposed BLGCN model and the corresponding HSI 
processing framework. First, we preprocess the input HSI raw data with the help 
of simple linear iterative cluster (SLIC) algorithm \cite{achanta2012slic}, and 
construct feature matrix and adjacency matrix. After that, we train our GAN 
network for the minority classes, augment the minority classes data, and update 
the adjacency matrix. 
We input the obtained features into the BLGCN model, and dynamically adjust 
the training strategy by quantifying the uncertainty of the output.

\subsection{Data Preprocessing}
In order to meet the input requirements of our model, we need to perform 
preprocessing operations on the original image. We first apply the SLIC 
algorithm to divide the original image into a number of 
superpixel blocks, the pixels in each block have strong spatial-spectral 
similarity with each other. The essence of the SLIC algorithm is $K$-means 
clustering, which starts to evolve from several initial points randomly placed 
on the image, and continuously clusters pixels with similar spatial-spectral 
characteristics. When the segmentation is completed, the original data can be 
simplified to a set of superpixel blocks.

For each superpixel block, we calculate the category distribution of pixels in 
the block and select the most occurring category as the label of the superpixel 
block. To improve the classification efficiency, we filter out all superpixel 
patches annotated as background. In order to extract the spatial information in 
the original image, we save the adjacency relationship between the superpixels 
in adjacency matrix and mark it as $\bold{A}$ based on the original spatial 
adjacency relationship between pixels.

We calculate the mean value of the pixel features contained over each dimension 
in superpixel block and save it as the texture feature $\bm{t}$, we sum the 
squares of the feature vector mean value and save it as the spectral feature 
$\bm{s}$. After concatenating the texture feature vector and the spectral 
feature vector we can get the feature matrix $\bm{F}=(\bm{t}\parallel\bm{s})$.

\subsection{Data Augmentation with GAN}
\begin{figure*}[th]
	\centering
	\includegraphics[scale=0.4]{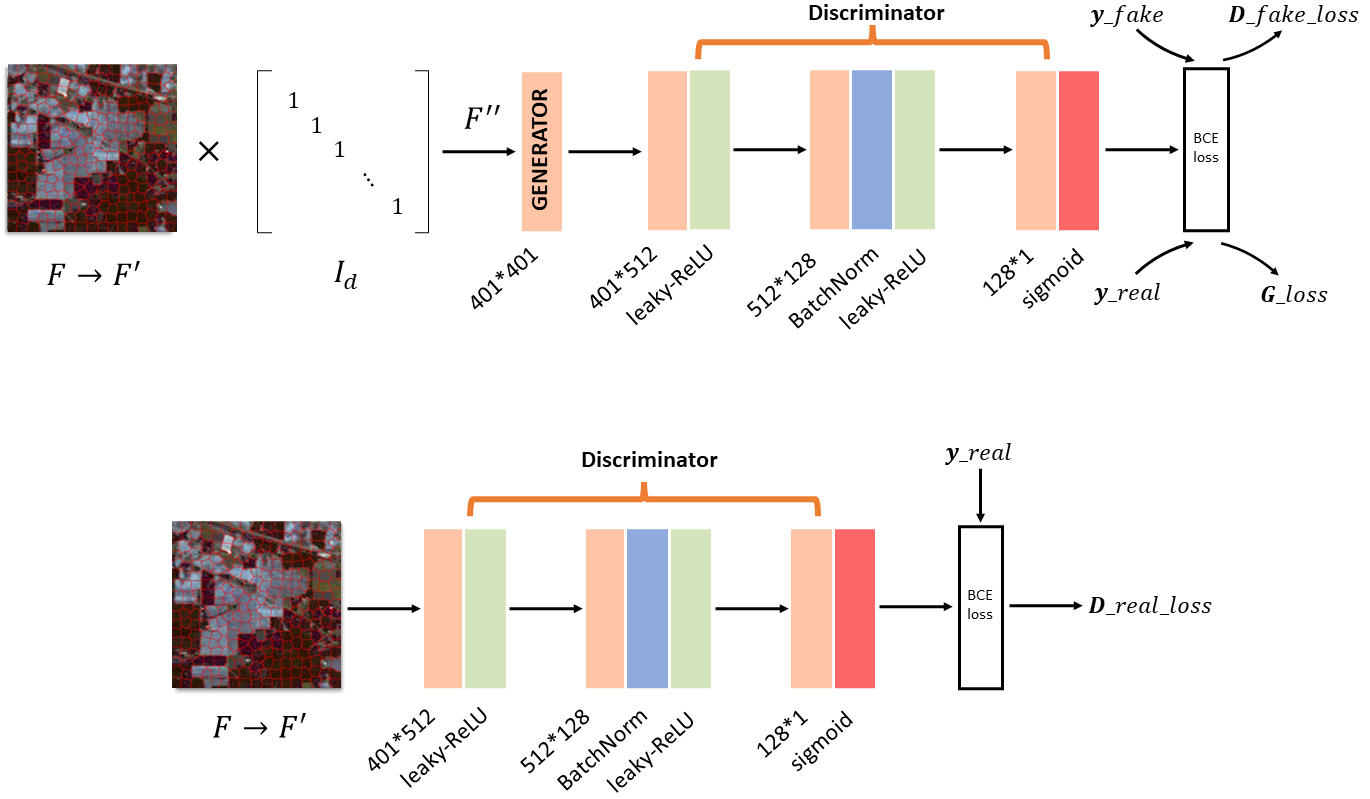}
	\caption{GAN-framework}
	\label{fig_framework}
\end{figure*}

Some of the preprocessed hyperspectral images still have a serious problem of 
sample imbalance. To solve it, we design a data augment framework based on 
the characteristics of feature matrix with the help of generative adversarial 
method, the framework is shown in Fig. \ref{fig_framework}, the integral 
settings in the framework are based on the Indian Pines HSI dataset.

\subsubsection[1.]{Formulation}
We design the generative adversarial data augmentation framework by answering 
two questions. The first one is how to increase the amounts of the minority 
class training samples inputted into the generative adversarial network while 
ensuring the quality of data, and the second task is how to learn and 
generate high-dimensional feature vectors without losing the original information.

\subsubsection[2.]{Feature Extraction}
After the image preprocessing process, for the sample class $m_i$, we mark the 
number of superpixel blocks in the class as $b_m$, and the feature vector of 
superpixel block as $f_m^j \in \mathbbm{R}^{(1 \times d)}, j \in [0,b_m)$, $d$ 
represents the dimension of the feature vector after preprocessing. We filter 
out the maximum amounts of superpixel blocks in the classes as 
$(b_m)_{max}$. For the class $m_i$ which meets the conditions of 
$b_{m_i}<0.02 \times (b_m)_{max}$, we define it as minority class, extract its 
feature vectors $f_{m_i}^j$ and splice it vertically into a minority class 
feature matrix $\bold{F}_i \in \mathbbm{R}^{b_{m_i }\times d}.$

Due to the high spectral dimension of the feature matrix, the existing 
mainstream spectral feature extraction methods and minority class data 
generation strategies are usually performed by applying PCA method 
\cite{shlens2014tutorial}, which leads to the loss of effective information. As 
a result, it reduces the classification accuracy of the model.

In order to achieve a balance between generation efficiency and data diversity, 
we design a transition matrix optimization strategy based on adversarial 
methods, we use the original feature vector as the input of the discriminator 
and train the discriminator's ability to identify minority class feature 
vectors. 
The feature matrix perturbed by the transition matrix is imported into the 
generator, then we input the generated result into the discriminator. The 
generator and discriminator are trained respectively. We assume that the weight 
matrix of the generator is $\bold{W}_{\bm{G}}$, in order to realize the 
perturbation operation on the original feature matrix, it is necessary to 
initialize the $mean(\bold{W}_{\bm{G}})=1$, and set a small variance to adapt 
to the data characteristic of the preprocessed feature matrix.

\subsubsection[3.]{Feature Enhancement}
For the purpose of cooperating with the optimization strategy designed, we 
construct a feature enhancement algorithm for minority class sequence data. For 
the minority class feature matrix $\bold{F}_i$ generated from preprocessing, we 
replicate it longitudinally to $d$-dimension and get $\bold{F}_i^{\prime}$, we 
import it into the discriminator as real data for training. We perform the 
operation $\bold{F}_i^{\prime\prime}=\bold{F}_i^{\prime}\times\bold{I}_d$ and 
do the 
perturbation to matrix $\bold{F}_i^{\prime\prime}$ with generator. Through this 
process, we construct a diagonal feature matrix that can be input into the 
generator for convolution, and the output of the generator has the 
characteristic that each row is a slice of new generated feature vector. By 
mixing in order, we effectively increase the data diversity of the training 
set. At the same time, if the replication order of the original feature matrix 
$\bold{F}_i$ to generate the feature matrix is adjusted, we will get different 
forms of enhanced features vector, which greatly reduces the overfitting 
probability of the model.

\subsubsection[4.]{Loss Function}

We design the loss function of the data augmentation model by inheriting the 
form of the GAN network loss function. To set an example, we only perform the 
data enhancement operation to a single class and ignore the mark of $i$, we 
define $\bm{y}_{real}$ and $\bm{y}_{fake}$ which have all elements equal to 1 
and 0 with a proper size. 
First we calculate the training loss $\bm{G}_{loss}$ of the generator 
\begin{equation}
\bm{G}_{loss}=BCE\_loss(\bm{D}(\bm{G}(\bold{F}^{\prime}\otimes 
\bold{I}),\bm{y}_{real}),
\end{equation} 
in which the $BCE\_loss$ refers to the binary cross entropy loss function, then 
input the original data and the generated data into the discriminator. Training 
loss $\bm{D}_{loss}$ with following equations:
\begin{align}
\bm{D}_{real\_loss}&=BCE\_loss(\bm{D}(\bold{F}),\bm{y}_{real})\\
\bm{D}_{fake\_loss}&=BCE\_loss(\bm{D}(\bm{G}(\bold{F}^{\prime}\otimes 
\bold{I}),\bm{y}_{fake})\\
\bm{D}_{loss}&=\bm{D}_{real\_loss}+\bm{D}_{fake\_loss}.
\end{align}

We define the final loss function as follows: 
\begin{equation}
\min_{\bm{G}}\max_{\bm{D}}V(\bm{D},\bm{G})=\bm{G}_{loss}-\bm{D}_{loss}.
\label{GAN_loss_function}
\end{equation}

After the training process, we take the output of the generator filling into 
the original feature matrix according to the model learning requirements, so as 
to realize the data enhancement for the minority class.

\subsection{Bayesian Layer and BLGCN}
\subsubsection{Bayesian Deep Learning}
In general, for a Bayesian deep neural network, its predicted value 
can be calculated by Eq. \ref{marginal probability}, and the calculation 
depends on the approximate inference of the posterior distribution $p(y^*\mid 
x^*,\bm{Y},\bm{X})$ expressed in Eq. \ref{Bayesian formula}. We 
take the approach of variational inference and use a simple distribution 
$q(\omega)$ to approximate the posterior distribution $p(\omega\mid 
\bm{Y},\bm{X}).$ 

In order to make $q(\omega)$ fit $p(\omega\mid \bm{Y},\bm{X})$ as closely as 
possible, we only need to minimize Kullback–Leibler divergence $KL[q(\omega) 
\parallel p(\omega\mid\bm{Y},\bm{X})]$ 
\cite{goan2020bayesian}.

To present uncertainty, we assume that both the approximate posterior and the 
prior follow Gaussian distribution, i.e.,
\begin{equation}
\begin{aligned}
q(\omega)=q(\omega\mid\theta),\theta=(\mu,\sigma)
\end{aligned}
\end{equation}
\begin{equation}
\begin{aligned}
p(\omega_i\mid\bm{Y},\bm{X})\sim\mathbbm{N}(\omega_i\mid\mu_i,{\sigma_i}^2).
\end{aligned}
\end{equation}

The original problem can be transformed into optimizing the parameter
\begin{equation}
\begin{aligned}
\theta^*&=\mathop{\arg\min}_{\theta}KL[q(\omega\mid\theta)\parallel 
p(\omega\mid\bm{Y},\bm{X}))] \\
&=\mathop{\arg\min}_{\theta}\mathbbm{E}_{q(\omega\mid\theta)}[\log\frac{q(\omega\mid\theta)}
{p(\omega\mid\bm{Y},\bm{X})}] \\
&=\mathop{\arg\min}_{\theta}\mathbbm{E}_{q(\omega\mid\theta)}
[\log\frac{q(\omega\mid\theta)p(\bm{Y}\mid\bm{X})}
{p(\omega\mid\bm{Y},\bm{X})p(\omega)}]\\
&=\mathop{\arg\min}_{\theta}\mathbbm{E}_{q(\omega\mid\theta)}[\log\frac{q(\omega\mid\theta)}
{p(\bm{Y}\mid\bm{X},\omega)p(\omega)}]
\end{aligned}
\end{equation}
and expressed as minimizing the loss function 
\begin{equation}
\begin{aligned}
Loss = KL[q(\omega\mid\theta)\parallel p(\omega)] - 
\mathbbm{E}_{q(\omega)}\log(p(\bm{Y}\mid\bm{X},\omega)) 
\label{Bayes_loss_function}.
\end{aligned}
\end{equation}

We apply the reparameter trick here \cite{shridhar2019comprehensive}, for 
the weights $\omega_i\sim\mathbbm{N}(\mu_i,{\theta_i}^2)$, we have 
$\omega_i=\mu_i+\theta_i\odot\epsilon_i$, where 
$\epsilon_i\sim\mathbbm{N}(0,1)$, and $\odot$ represents Hadamard product. Then 
for the backpropagation of the loss function Eq. \ref{Bayes_loss_function}, we 
have 
\begin{equation}
\begin{aligned}
\frac{\partial}{\partial\theta}\mathbbm{E}_{q(\epsilon)}
\{\log[\frac{q(\omega\mid\theta)}{p(\bm{Y}\mid\bm{X},\omega)p(\omega)}]\} \\
=\mathbbm{E}_{q(\epsilon)}\{\frac{\partial}{\partial\theta}
\log[\frac{q(\omega\mid\theta)}{p(\bm{Y}\mid\bm{X},\omega)p(\omega)}]\}.
\end{aligned}
\end{equation}

In order to ensure that the range of parameter $\theta$ includes the entire 
real number axis, we need to perform the reparameter trick on it, with 
$\sigma=\log(1+e^{\rho})$ we have $\theta=(\mu,\rho).$
\begin{figure}[t]
	\includegraphics[width=\linewidth]{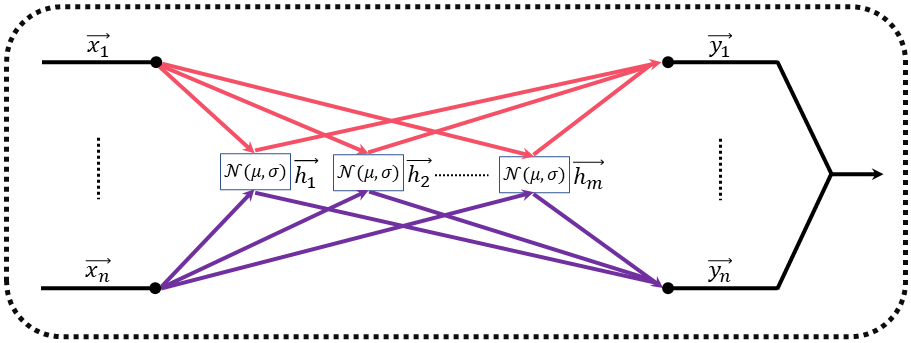}
	\caption{Bayesian Layer} 
	\label{BayesianLayer.png}
\end{figure}
\subsubsection{Bayesian Layer}

The Bayesian neural network used in the current research field is relatively 
fixed in form, and has not been effectively combined with the convolutional 
neural network \cite{shridhar2019comprehensive}. Researchers usually simply 
migrate the Bayesian forward propagation method and loss function to the 
traditional neural networks, and ignore adjusting the network structure in 
combination with specific tasks. Since the weight matrix of Bayesian neural 
network obeys a specific distribution, whether it is dealt with variable 
inference method or the MCMC method, we need to perform multiple sampling in 
the forward propagation with approximate calculation, which will take a great 
time cost. And when dealing with the hyperspectral image classification task,  
directly applying Bayesian theory to completely transform graph convolutional 
neural networks limits the room to fine-tune the network framework and improve 
the classification results.

Therefore, we propose the concept of Bayesian layer. We define the Bayesian 
layer as a convolutional layer which all weight parameters and biases are 
represented in distributed form as shown in Fig. \ref{BayesianLayer.png}. In 
the process of forward propagation, the input features matrix will be divided 
into several row vectors $\overrightarrow{x_n}$, and the Bayesian matrix can be 
seen as a set of column vectors $\overrightarrow{h_n}$, which is assembled by 
weight parameters of $\mu$ and $\theta$. In the matrix multiplication 
operation, we have $\overrightarrow{y_{nm}} = \overrightarrow{x_n} \times 
\overrightarrow{h_m}$. 

At the same time, we stipulate that the loss function must have the form of KL 
divergence which can be optimized by the variational inference method. By 
analyzing the form of the Bayesian convolutional neural network loss function, 
we find that the likelihood function term can be replaced by a 
multi-classification loss. The reason is, given the model output $y$, the 
likelihood function $L(\omega\mid y)$ about the parameter $\omega$ is 
numerically equals to the probability $p(Y=y\mid\omega)$ of the output $y$ 
given the parameter $\omega$. In the process of each forward propagation, the 
weight parameters of the non-Bayesian layer remain unchanged when the Bayesian 
layer is sampled multiple times, so each time the gradient is updated in the 
backpropagation, the existence of non-Bayesian layer will not influence the 
derivation operation towards the Bayesian layer.

As a result, if we use the classification loss function of traditional neural 
network such as negative log-likelihood loss (nll loss) to represent the 
likelihood function, it can provide the theoretical basis for the realization 
of the Bayesian layer. Here we can rewrite Eq. \ref{Bayes_loss_function} as 
follows:

\begin{equation}
\begin{aligned}
\begin{split}
Loss &= \sum_{i}\log{q(\omega_{i}\mid\theta_{i})} - 
\sum_{i}\log{p(\omega_{i})}\\ 
&-\sum_{j}\log{p(y_{j}\mid\omega,x_{j})} \\
&= \sum_{i}\log{q(\omega_{i}\mid\theta_{i})} - \sum_{i}\log{p(\omega_{i})} - 
nll\_loss.
\label{loss_function}
\end{split}
\end{aligned}
\end{equation}

According to the former demonstration, when upgrading the weight parameters of 
a Bayesian layers convolutional neural network in gradient calculation and 
back propagation, we know that the prior distribution loss 
$\sum_{i}\log{p(\omega_{i})}$ and posterior distribution loss 
$\sum_{i}\log{q(\omega_{i}\mid\theta_{i})}$ only relate to the weights of 
Bayesian layer, and the multi-classification loss relates to all the weight 
parameters of the neural network. Therefore, the parameter updates of the 
Bayesian layer and the non-Bayesian layer can be performed sequentially under 
the same process without interfering with each other.

Meanwhile, it also gives us more choices in diversification. Facing 
various tasks, we can choose different insertion methods of the Bayesian 
layer, which can achieve the uncertainty quantization output while ensuring 
that the loss function structure of the original neural network is not damaged.

\subsubsection{BLGCN}
We combine the idea of graph convolution with Bayesian layers, where the output 
of each Bayesian layer is multiplied with the preprocessed adjacency matrix to 
fuse the spatial features of hyperspectral images. 
More specifically, the adjacency matrix $\bold{A}$ got from the preprocessing 
is renormalized with the method proposed by Kipf\&Welling \cite{kipf2016semi}. 
To generalize the formulation of $\hat{\bold{A}}_G$, we perform the trick 
\begin{equation}
\begin{aligned}
\hat{\bold{A}}_G = 
\hat{\bold{D}}^{-\frac{1}{2}}\hat{\bold{A}}\hat{\bold{D}}^{-\frac{1}{2}} 
\end{aligned}
\end{equation}
with $\hat{\bold{A}} = \bold{A} + \bold{I}$ and $\hat{\bold{D}}_{ii} = 
\sum\nolimits_{j}\hat{\bold{A}}_{ij}$. Here, $\bold{I}$ denotes the identity 
matrix which has proper size to calculate. Hence, for each Bayesian layer, the 
output can be expressed with Eq. \ref{GCN} after sampling on the weight 
parameter matrix.

\begin{figure*}
	\centering
	\includegraphics[scale=0.4]{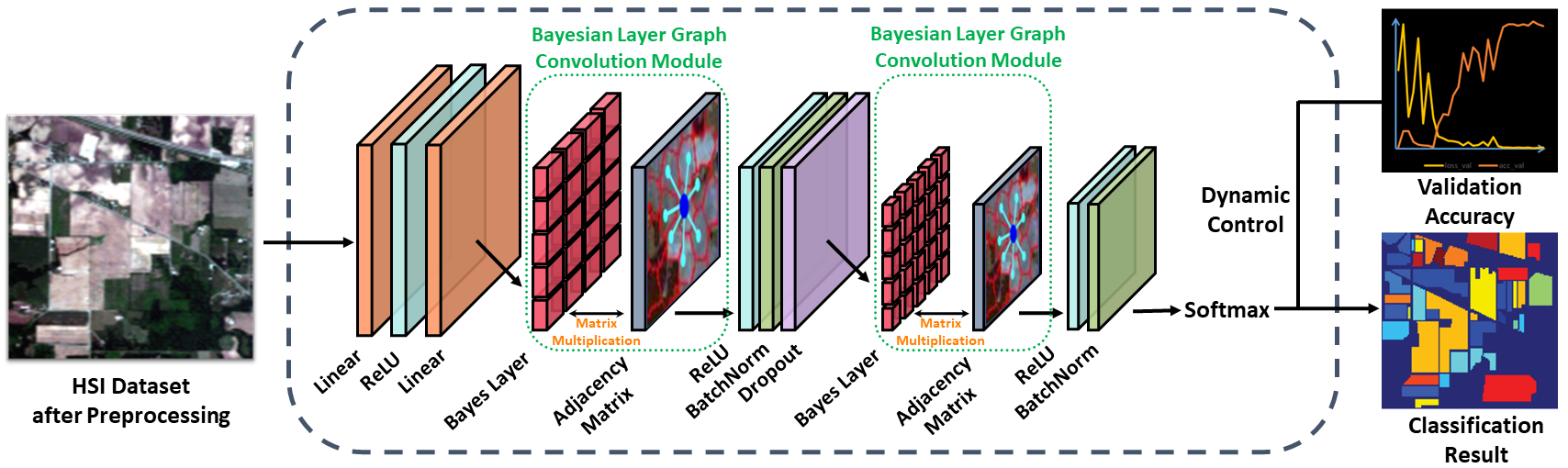}
	\caption{Bayesian layer graph convolutional network}
	\label{fig_NerualNetwork}
\end{figure*}

The main framework of BLGCN is shown in Fig. \ref{fig_NerualNetwork}, we set a 
feature extraction module before the two Bayesian layer modules, which consists 
of two full-connected convolutional layers and a ReLU layer, it can roughly 
extract the data features and provide a solid foundation for the Bayesian 
processing.

It should be noted that for the data-enhanced hyperspectral data, in order to 
maintain the consistency of the spatial information, the adjacency matrix 
needs to be expanded. We assume that the newly generated superpixels have 
similar spatial characteristics with the original superpixels of the same 
class. We randomly match the generated ones with those initial superpixels, and 
give them the same adjacency relationship with the matched ones, thereby 
regenerating a new adjacency matrix.
\subsection{Training Strategy based on Bayesian Method}
The prediction results given by deep learning models are not always reliable, 
and some application fields of hyperspectral image classification have high 
requirements for confidence. By modeling uncertainty with output variance, we 
dynamically evaluate the confidence level of the training results and make 
decisions on whether terminate the training process.

From the central limit theorem, we know that if the sample size is large 
enough, the sampling distribution can be approximated as a normal distribution. 
Due to the existence of the Bayesian layer, we can consider that outputs of the 
multiple times forward propagation after fixing the model are independent from 
each other. Therefore, by calculating the mean and variance of the multiple 
forward propagation output, we can obtain an interval estimate of the accuracy 
and a confidence interval under the premise of given confidence.

To enable dynamic evaluation of the training process, we set two accuracy 
thresholds and calculate the accuracy of the validation set after each training 
process. When it reaches the first threshold, the model weights are fixed after 
each training batch and the forward propagation is performed 30 times. We 
record the output results and calculate whether the confidence interval upper 
bound of the validation set accuracy reaches the second threshold we set under 
95\% confidence level, and if it does, the training will be stopped 
immediately. For training tasks with specific sample classification accuracy 
requirements, we only need to change the two judgment operations on the 
validation set accuracy to the specific class accuracy.

We applied statistical knowledge to calculate the confidence interval of the 
accuracy. For the case of large sample size, we used the Z-test to determine 
whether the sample mean was significantly different from the population mean. 
We first calculate the sample mean $\mu$ and standard error $SE$, then 
determine the confidence level and get the standard score $z$. When the 
confidence level is set to 95\%, the standard score $z$ is numerically equal 
to 1.96, and the bounds of the confidence interval $[a,b]$ are shown in below:
\begin{equation}
\begin{aligned}
a = \mu - \left|z\right| \times SE \\
b = \mu + \left|z\right| \times SE.
\label{SE}
\end{aligned}
\end{equation}

We combine the proposed model with dynamic control strategy, and formulate the 
whole classification process in forms of flow chart. The implementation details 
of the process are shown in Algorithm~\ref{alg:algorithm1}.
\begin{algorithm}[th]
	\caption{Proposed BLGCN with Dynamic Control Strategy for HSI 
	Classification} 
	\label{alg:algorithm1}
	\hspace*{0.02in} {$\bold{Input:}$} 
	Input image; number of iterations $i$; Threshold $\bm{T_1}$,$\bm{T_2}$\\
	\hspace*{0.02in} {$\bold{Output:}$} 
	Classification Result
	\begin{algorithmic}[1]
		\State // Preprocessing the dataset
		\State Segment the input image into superpixels with SLIC;
		\State Label the superpixels and exclude the background;
		\State Construct the adjacency matrices $\bm{A}$;
		\State Extract the image information into the feature matrix $\bm{F}$;
		\If{The sample imbalance problem is serious} 
			\State Train the GAN network on preprocessed dataset with Eq. 
			\ref{GAN_loss_function};
			\State Expand the feature matrix $\bm{F}$ with generated data;
			\State Update the adjacency matrices $\bm{A}$;
		\EndIf
		\State // Training process with dynamic control
		\For{epoch in range $i$} 
			\State Train the model with BLGCN with Eq. \ref{loss_function};
			\If{validation set accuracy reaches $\bm{T_1}$} 
				\State Fix the model;
				\State Perform the forward propagation 30 times;
				\State Calculate the confidence interval with Eq. \ref{SE};
				\If{confidence interval reaches $\bm{T_2}$}
				\State \textbf{Break}
				\EndIf
			\EndIf
		\EndFor
		\State \Return result.
	\end{algorithmic}
\end{algorithm}

\section{Experimental Analysis}
\subsection{Dataset Description and Evaluation Criteria}

In our experiment, we select four mainstream hyperspectral remote sensing datasets, including 
Indian Pines, Salinas, Pavia University and Houston University to verify the advantages of the 
model we proposed.

\begin{figure}[htbp]
	\includegraphics[width=\linewidth]{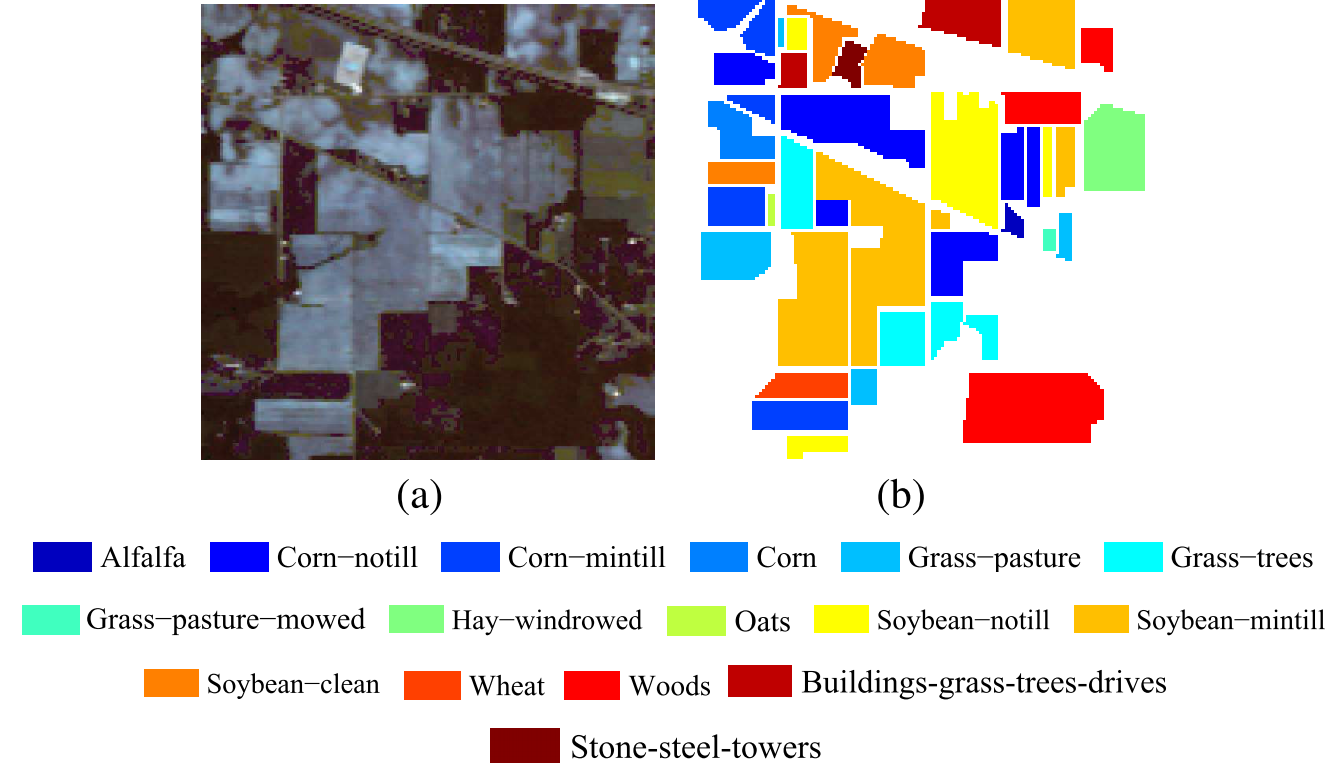}
	\caption{Indian Pines. (a) False color image. (b) Ground truth map.} \label{indian.png}
\end{figure}

Indian Pines dataset was obtained by AirBorne Visible\&Infrared Imaging 
Spectrometer sensor over the Indian Pines region in 1992. It contains 
145$\times$145 pixels and consists of 220 spectral bands covering the range 
from 0.4 $\mu$m to 2.5 $\mu$m. 20 bands absorbed by water vapor are removed 
and the remaining 200 spectral bands are used for model classification. This 
dataset contains 16 kinds of landscapes. After the image preprocessing, the 
pixels in the dataset are aggregated into several superpixels. The number of 
superpixels in each landscape class and their corresponding 
training and testing samples are listed in Table \ref{Table1}. The 
ground-truth map is shown in Fig. \ref{indian.png}.

\begin{table}[htbp]\footnotesize 
	\renewcommand\arraystretch{1}
	\centering  
	\caption{\\NUMBERS OF LABELED AND UNLABELED SUPERPIXELS USED IN THE INDIAN 
		PINES DATASET}  
	\label{Table1}  
	
		\begin{tabular}{ccccc} 
			\toprule 
			ID  & & Class                        & Labeled & Unlabeled 
			\\
			\midrule 
			1   & & Alfalfa                      & 5   &  41 \\    
			2   & & Corn-notill                  & 143 &  1285 \\
			3   & & Corn-mintill                 & 83 &   747\\
			4   & & Corn                         & 24 &   213\\
			5   & & Grass-pasture                & 48 &   435\\
			6   & & Grass-trees                  & 73 &   657\\
			7   & & Grass-pasture-mowed          & 3 &   25\\
			8   & & Hay-windrowed                & 48 &  430 \\
			9   & & Oats                         & 2 &   18\\
			10  & & Soybean-notill               & 97 &  875 \\
			11  & & Soybean-mintill              & 246 & 2209  \\
			12  & & Soybean-clean                & 59 &  534 \\
			13  & & Wheat                        & 21 &  185 \\
			14  & & Woods                        & 127 & 1138  \\
			15  & & Buildings-grass-trees-drives & 39 &  347 \\
			16  & & Stone-steel-towers           & 9 &  84 \\
			\bottomrule 
		\end{tabular}
	
\end{table}

\begin{figure}[htbp]
	\includegraphics[width=\linewidth]{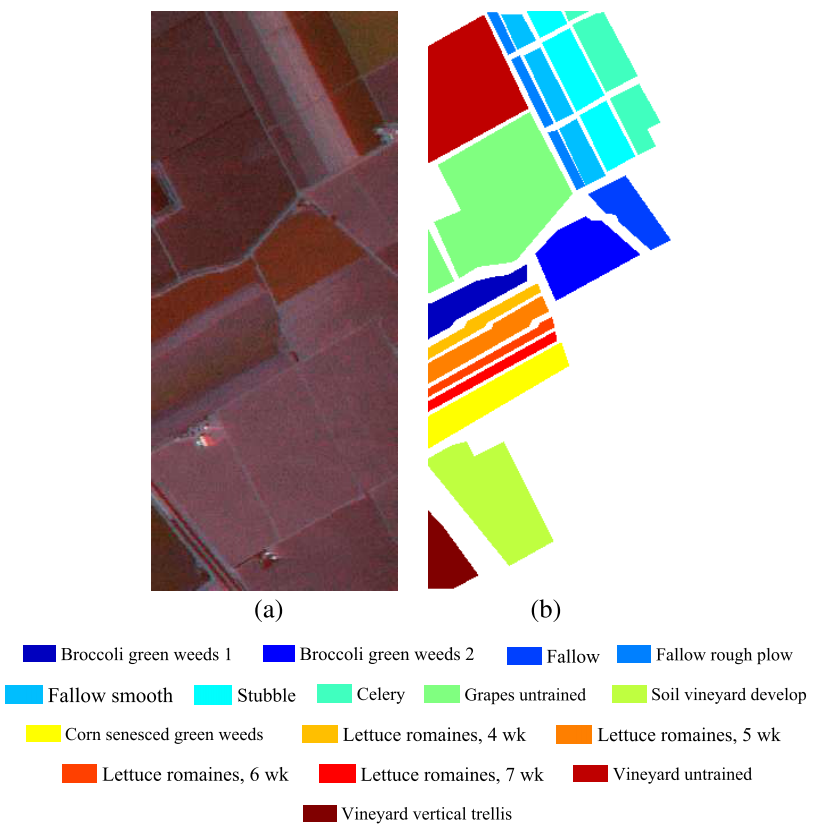}
	\caption{Salinas. (a) False color image. (b) Ground truth map.} 
	\label{salinas.png}
\end{figure}

The University of Pavia dataset was acquired over the Pavia University in Italy 
in 2001. It contains 610$\times$340 pixels in 103 spectral bands and has a 
wavelength range from 0.43 $\mu$m to 0.86 $\mu$m after removing the noisy 
bands. 
Table \ref{Table3} lists the details of the superpixels and the amounts of 
training and testing samples. Fig. \ref{salinas.png} shows the ground-truth 
map of the dataset.
\begin{figure}[htbp]
	\includegraphics[width=\linewidth]{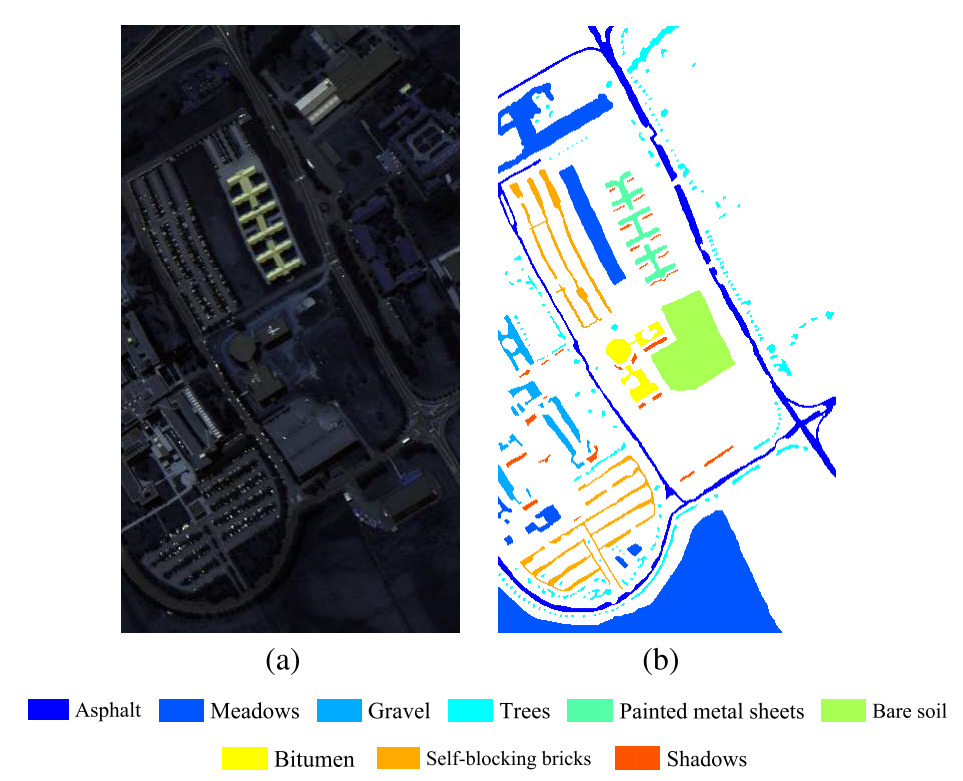}
	\caption{Pavia University. (a) False color image. (b) Ground truth map.} 
	\label{paviau.png}
\end{figure}

\begin{table}[hptb]\small 
	\renewcommand\arraystretch{1}
	\centering  
	\caption{\\NUMBERS OF LABELED AND UNLABELED SUPERPIXELS USED IN THE PAVIA 
		UNIVERSITY}  
	\label{Table2}  
	\setlength{\tabcolsep}{2.65mm}{
		\begin{tabular}{ccccc} 
			\toprule 
			ID  & & Class                & Labeled & Unlabeled 
			\\
			\midrule 
			1   & & Asphalt              & 663 & 5968  \\    
			2   & & Meadows              & 1854 & 16795  \\
			3   & & Gravel               & 210 & 1889  \\
			4   & & Trees                & 306 & 2758  \\
			5   & & Painted metal sheets & 135 & 1210  \\
			6   & & Bare soil            & 503 & 4526  \\
			7   & & Bitumem              & 133 & 1197  \\
			8   & & Self-blocking bricks & 368 & 3314  \\
			9   & & Shadows              & 95 &  852 \\
			\bottomrule 
		\end{tabular}
	}
\end{table}

Salinas dataset captured the Salinas Valley in California. This dataset 
contains 512$\times$127 pixels in 224 bands. 20 bands absorbed by water vapor 
are removed and the remaining 204 spectral bands are used for classification. 
Its ground-truth map is shown in Fig. \ref{paviau.png} and the number of 
superpixels with their information are listed in Table \ref{Table2}.

\begin{table}[hptb]\small 
	\renewcommand\arraystretch{1}
	\centering  
	\caption{\\NUMBERS OF LABELED AND UNLABELED SUPERPIXELS USED IN THE SALINAS 
		DATASET}  
	\label{Table3}  
	
		\begin{tabular}{ccccc} 
			\toprule 
			ID  & & Class                      & Labeled & Unlabeled 
			\\    
			\midrule 
			1   & & Brocoli-green-weeds-1      & 201 & 1808  \\    
			2   & & Brocoli-green-weeds-2      & 373 & 3353  \\
			3   & & Fallow                     & 198 & 1778  \\
			4   & & Fallow-rough-plow          & 139 & 1255  \\
			5   & & Fallow-smooth              & 268 & 2410  \\
			6   & & Stubble                    & 393 & 3536  \\
			7   & & Celery                     & 358 & 3221   \\
			8   & & Grapes-untrained           & 1127 & 10144  \\
			9   & & Soil-vinyard-develope      & 620 & 5583  \\
			10  & & Corn-senesced-green-weeds  & 328 & 2950  \\
			11  & & Lettuce-romaine-4wk        & 107 & 961  \\
			12  & & Lettuce-romaine-5wk        & 193 & 1734  \\
			13  & & Lettuce-romaine-6wk        & 92 & 824  \\
			14  & & Lettuce-romaine-7wk        & 107 & 963  \\
			15  & & Vinyard-untrained          & 727 & 6541  \\
			16  & & Vinyard-vertical-trellis   & 181 & 1626  \\
			\bottomrule 
		\end{tabular}
	
\end{table}

\begin{figure}[htbp]
	\includegraphics[width=\linewidth]{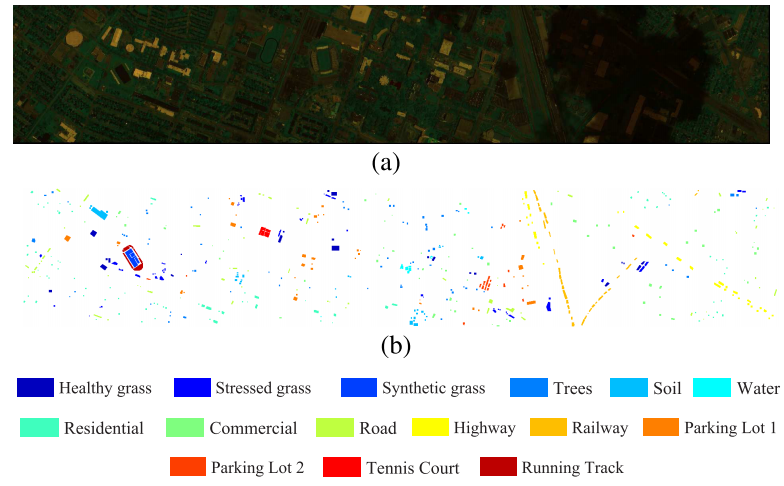}
	\caption{Houston University. (a) False color image. (b) Ground truth map.} 
	\label{houstonu.png}
\end{figure}

The University of Houston dataset is provided by the IEEE GRSS Data Fusion 
Contest in 2013. This dataset consists of 349$\times$1905 pixels and contains 
144 spectral bands covering the range from 364 nm to 1046 nm. The number of 
superpixels in each landscape class and their corresponding training and 
testing samples are given in Table \ref{Table4} with its ground-truth map shown 
in Fig. \ref{houstonu.png}.

\begin{table}[hptb]\small 
	\renewcommand\arraystretch{1}
	\centering  
	\caption{\\NUMBERS OF LABELED AND UNLABELED SUPERPIXELS USED IN THE HOUSTON 
		UNIVERSITY}  
	\label{Table4}
	\setlength{\tabcolsep}{3.5mm}{
		\begin{tabular}{ccccc} 
			\toprule 
			ID  & & Class   \qquad     & Labeled & Unlabeled 
			\\ 
			\midrule 
			1   & & Healthy grass      & 125 & 1126  \\    
			2   & & Stressed grass     & 125 & 1129  \\
			3   & & Synthetic grass    & 70 &  627 \\
			4   & & Trees              & 124 & 1120  \\
			5   & & Soil               & 124 & 1118  \\
			6   & & Water              & 33 & 292  \\
			7   & & Residential        & 127 & 1141  \\
			8   & & Commercial         & 124 & 1120  \\
			9   & & Road               & 125 & 1127  \\
			10  & & Highway            & 123    & 1104     \\
			11  & & Railway            & 124    & 1111     \\
			12  & & Parking Lot1       & 123    & 1110      \\
			13  & & Parking Lot2       & 47    &  422      \\
			14  & & Tennis Court       & 43    &  385     \\
			15  & & Running track      & 66    &  594     \\
			\bottomrule 
		\end{tabular}
	}
\end{table}

In our experiment, four metrics are adopted to quantificationally evaluate the 
classification performance which are per-class accuracy, overall accuracy (OA), 
average accuracy (AA), and kappa coefficient (Kappa).

\subsection{Experimental Settings}
In our experiment, the proposed model is implemented via Pytorch with the Adam 
optimizer. We train our model with 0.2 dropout rate, 5$\times 10^{-4}$ weight 
decay rate. We adopt the learning rate decay strategy in the training 
process. We set the initial learning rate as 1$\times10^{-3}$ and use the 
multistep learning rate strategy with gamma fixed to 0.9, and three milestones 
are 1500, 2500 and 3500. The threshold parameter of assigning the pseudo label 
to the unlabeled nodes is set to 0.9, which means that besides the training 
sample, all the rest samples are distributed to the validation set.

In the SLIC algorithm in image preprocessing, we set the compactness parameter 
to 0.08 for all experiments. And in our GAN algorithm in data augment process, 
the initial standard deviation of the generator weight matrix and the 
discriminator weight matrix are set to 1$\times10^{-5}$ and 0.01, respectively. 
Because of the data characteristics after the normalization operation in data 
preprocessing, the learning rate in our GAN network is set to 
1${\times10^{-7}}$. 
For the landscape class  which has the superpixels sample size less than 0.02 
times of the superpixels sample size of the largest class sample, we define it 
as minority class and perform the data augment process towards it, 
the generated samples are filled into the original dataset until the sample 
size of the minority class reaches the threshold of 0.05 times of the 
superpixels 
sample size of the largest class sample.

\subsection{Classification Results}
In order to evaluate the effectiveness of our proposed BLGCN model, we conduct 
comparative experiments with various comparison algorithms on the selected 
dataset. To make the experiment results reliable, various SOTA algorithms 
are selected as the competitors. Among the algorithms based on traditional 
machine learning, we selected the support vector machine algorithm (SVM) 
\cite{kuo2010spatial}. The commonly used SVM kernel includes linear kernel, 
polynomial kernel, Gaussian kernel and Sigmoid kernel. We make 
the SVM network search for optimization in the given parameter set, and then 
select the appropriate hyperparameters for the classification task. For 
convolutional neural network based on deep learning methods, we select the 
3D-CNN convolutional neural network \cite{hamida20183} which performs well over 
the HSI datasets. For graph convolutional neural network based model, we select 
several algorithms that have been widely used in recent years, such as GCN 
\cite{kipf2016semi}, spectral-spatial graph convolutional network (S$^2$GCN) 
\cite{qin2018spectral}, and multiscale dynamic GCN (MDGCN) 
\cite{wan2019multiscale}, etc. They are powered by the GCN idea to improve the 
network in a targeted manner, and have achieved better classification results 
on the HSI datasets. We perform 30 independent experiments on each dataset 
based on the algorithm chosen based on the application and recorded the mean 
and standard deviation of the results. It is worth noting that since the 
classification accuracies are quite poor when the 3D-CNN model is performed on 
the HSI datasets with 10\% training set ratio, we only show the classification 
results with 30\% training set ratio on the last three datasets.

The classification results on the Indian Pines dataset show the effectiveness 
of our algorithm. The traditional algorithm (SVM) and the original GCN have 
serious misclassification problem in the face of minority class categories 
classification task, and the accuracy of the 3D-CNN method based on 
full-supervised classification is also relatively low. When the proportion of 
training set samples is increased from 10\% to 30\%, the classification effect 
has been significantly improved. Compared with other GCN methods, our method 
has obvious advantages in the task of distinguishing minority class categories, 
i.e., Class 7 and Class 9, while maintaining high classification accuracy on 
other sample categories. The relevant results are listed in Table \ref{Table5}.
\begin{table*}[hptb]\normalsize 
	\renewcommand\arraystretch{1}
	\centering  
	\caption{\\CLASSIFICATION RESULTS ON INDIAN PINES DATASET}  
	\label{Table5}
	\resizebox{0.8\linewidth}{!}{
		\begin{tabular}{ccccccccc} 
			\toprule 
			Class  & & SVM & 3D-CNN(0.1) & 3D-CNN(0.3) & GCN & S$^2$GCN & MDGCN 
			& BLGCN
			\\
			\midrule 
			1   & & 59.81$\pm$11.20 & 78.93$\pm$16.75 & 
			\pmb{95.55}$\pm$\pmb{3.99} & 
			73.27$\pm$16.47 & 92.05$\pm$5.64 & 90.02$\pm$4.62 & 
			92.30$\pm$3.63\\    
			2   & & 73.53$\pm$1.70  & 82.19$\pm$1.75  & 94.42$\pm$1.43 & 
			81.03$\pm$1.66 & 82.84$\pm$2.32 & 77.83$\pm$1.87 & 
			\pmb{96.80}$\pm$\pmb{1.19}\\
			3   & & 65.63$\pm$2.81  & 77.25$\pm$2.98  & 89.73$\pm$1.49 & 
			81.90$\pm$0.79 & 80.34$\pm$3.15 & 95.39$\pm$2.56 & 
			\pmb{95.94}$\pm$\pmb{3.27}\\
			4   & & 54.66$\pm$5.29  & 76.59$\pm$3.90  & 92.81$\pm$0.99 & 
			69.76$\pm$0.64 & 90.10$\pm$1.85 & 95.69$\pm$1.49 & 
			\pmb{96.92}$\pm$\pmb{6.19}\\
			5   & & 88.67$\pm$1.77  & 92.41$\pm$1.82  & 
			\pmb{95.69}$\pm$\pmb{0.87} & 
			95.44$\pm$0.42 & 94.10$\pm$0.1  & 92.36$\pm$0.04 & 93.25$\pm$4.50\\
			6   & & 92.48$\pm$1.07  & 97.41$\pm$0.82  & 99.09$\pm$0.23 & 
			\pmb{100.00}$\pm$\pmb{0.00}& 94.25$\pm$0.24 & 94.32$\pm$0.16 & 
			99.40$\pm$0.80\\
			7   & & 75.42$\pm$7.90  & 80.53$\pm$12.10 & 92.75$\pm$3.29 & 
			0.00$\pm$0.00 & 67.49$\pm$6.22 & 82.03$\pm$5.09 & 
			\pmb{93.30}$\pm$\pmb{7.90}\\
			8   & & 96.22$\pm$0.90  & 98.41$\pm$0.89  & 99.86$\pm$0.14 & 
			\pmb{100.00}$\pm$\pmb{0.00}& 95.93$\pm$0.03 & 96.00$\pm$0.00 & 
			\pmb{100.00}$\pm$\pmb{0.00}\\
			9   & & 36.98$\pm$13.95 & 73.95$\pm$15.89 & 95.62$\pm$3.80 & 
			0.00$\pm$0.00 
			& 75.13$\pm$7.85 & 79.02$\pm$6.44 & \pmb{100.00}$\pm$\pmb{0.00}\\
			10  & & 69.44$\pm$1.92  & 83.21$\pm$2.03  & 
			\pmb{93.02}$\pm$\pmb{1.27} & 
			79.26$\pm$0.94 & 90.86$\pm$0.95 & 87.39$\pm$0.74 & 92.19$\pm$4.00\\
			11  & & 77.65$\pm$1.43  & 85.97$\pm$1.61  & 94.69$\pm$0.76 & 
			91.73$\pm$0.08 & 82.38$\pm$2.14 & 90.62$\pm$1.72 & 
			\pmb{98.08}$\pm$\pmb{1.21}\\
			12  & & 70.43$\pm$4.76  & 80.17$\pm$3.01  & 93.77$\pm$1.11 & 
			86.66$\pm$1.20 & 77.62$\pm$4.79 & 90.33$\pm$3.91 & 
			\pmb{97.94}$\pm$\pmb{3.29}\\
			13  & & 94.12$\pm$2.11  & 97.57$\pm$1.51  & 99.29$\pm$0.44 & 
			93.14$\pm$2.80 & 96.32$\pm$0.72 & 97.08$\pm$0.55 & 
			\pmb{99.86}$\pm$\pmb{0.33}\\
			14  & & 93.85$\pm$0.71  & 96.15$\pm$0.95  & 98.73$\pm$0.34 & 
			97.37$\pm$0.98 & 93.71$\pm$1.45 & 96.81$\pm$1.15 & 
			\pmb{99.72}$\pm$\pmb{0.82}\\
			15  & & 59.32$\pm$3.54  & 73.52$\pm$3.76  & 87.91$\pm$1.44 & 
			86.86$\pm$0.60 & 84.13$\pm$1.18 & \pmb{96.81}$\pm$\pmb{0.93} & 
			95.07$\pm$3.12\\
			16  & & 94.34$\pm$2.33  & 96.64$\pm$2.13  & 99.00$\pm$0.62 & 
			96.14$\pm$1.70 & 96.87$\pm$1.18 & 92.91$\pm$0.00 & 
			\pmb{100.00}$\pm$\pmb{0.00}\\
			\midrule
			OA  & & 78.78$\pm$1.03  & 86.08$\pm$0.77  & 94.24$\pm$0.58 & 
			88.80$\pm$0.53
			& 84.91$\pm$1.79 & 87.09$\pm$1.33 & \pmb{97.22}$\pm$\pmb{0.81}\\
			AA  & & 75.09$\pm$3.93  & 85.62$\pm$2.38  & 95.26$\pm$0.44 & 
			75.97$\pm$0.45
			& 87.08$\pm$2.23 & 90.25$\pm$1.98 & \pmb{97.02}$\pm$\pmb{1.15}\\
			Kappa&& 75.70$\pm$1.20  & 84.14$\pm$0.87  & 93.46$\pm$0.68 & 
			87.21$\pm$0.57
			& 83.02$\pm$0.98 & 86.53$\pm$0.86 & \pmb{96.81}$\pm$\pmb{0.94}\\
			\bottomrule 
		\end{tabular}
	}
\end{table*}

Table \ref{Table6} shows the classification effects of different algorithms on 
the Pavia University dataset. Among these methods, the algorithm we proposed 
still maintains the highest classification accuracy, and has achieved 100\% 
recognition and classification accuracy in several categories. It is 
worth noting that due to the relatively balanced sample size of the PU dataset, 
the 3D-CNN model also reaches a good classification effect after we increase 
the training set ratio to 30\%, and gets the highest classification accuracy in 
some categories. For the samples class that are scattered in spatial such as 
Class 4, the classification methods based on GCN fail to achieve good results. 
The main reason is that its spatial adjacency relationship is relatively 
sparse, and makes it easy to be misclassified when using the GCN methods.
\begin{table*}[hptb]\scriptsize 
	\renewcommand\arraystretch{1}
	\centering  
	\caption{\\CLASSIFICATION RESULTS ON PAVIA UNIVERSITY DATASET}  
	\label{Table6}
	\resizebox{0.8\linewidth}{!}{
		\begin{tabular}{cccccccc} 
			\toprule 
			Class  & & SVM & 3D-CNN(0.3) & GCN & S$^2$GCN & MDGCN & BLGCN\\
			\midrule
			1 & & 92.40$\pm$1.71 & \pmb{97.94}$\pm$\pmb{0.10} & 70.97$\pm$0.57 
			& 90.69$\pm$1.84 & 
			91.09$\pm$0.45 &
			96.45$\pm$4.34\\
			2 & & 95.54$\pm$0.84 & 97.44$\pm$0.05 & 96.55$\pm$0.28 & 
			85.02$\pm$3.22 & 96.64$\pm$0.41 &
			\pmb{99.12}$\pm$\pmb{0.75}\\
			3 & & 79.92$\pm$1.23 & \pmb{94.56}$\pm$\pmb{0.26} & 24.74$\pm$0.55 
			& 85.91$\pm$3.18 & 
			89.61$\pm$0.29 &
			83.25$\pm$11.57\\
			4 & & 93.70$\pm$2.17 & \pmb{98.02}$\pm$\pmb{0.13} & 47.25$\pm$1.61 
			& 88.72$\pm$1.24 & 
			81.57$\pm$1.97 &
			66.76$\pm$1.49\\
			5 & & 99.55$\pm$0.22 & 99.88$\pm$0.04 & 100.00$\pm$0.00& 
			97.66$\pm$0.98 & 96.86$\pm$0.21 &
			\pmb{100.00}$\pm$\pmb{0.00}\\
			6 & & 85.61$\pm$2.29 & 98.26$\pm$0.19 & 99.95$\pm$0.15 & 
			86.52$\pm$1.59 & 92.75$\pm$0.48 &
			\pmb{100.00}$\pm$\pmb{0.00}\\
			7 & & 76.53$\pm$8.90 & \pmb{98.36}$\pm$\pmb{0.09} & 94.24$\pm$1.39 
			& 96.57$\pm$0.76 & 
			96.32$\pm$1.26 &
			98.00$\pm$2.58\\
			8 & & 86.22$\pm$1.10 & 96.52$\pm$0.37 & 
			\pmb{100.00}$\pm$\pmb{0.00}& 87.86$\pm$3.01 & 
			92.49$\pm$1.00 &
			91.71$\pm$1.83\\
			9 & & 99.90$\pm$1.41 & 99.68$\pm$0.11 & 0.00$\pm$0.00  & 
			96.58$\pm$0.79 & 78.89$\pm$0.78 &
			\pmb{100.00}$\pm$\pmb{0.00}\\
			\midrule
			OA& & 91.96$\pm$1.41 & 96.58$\pm$0.04 & 90.84$\pm$0.19 & 
			87.64$\pm$1.82 & 93.16$\pm$0.26 &
			\pmb{97.56}$\pm$\pmb{0.38}\\
			AA& & 89.82$\pm$0.28 & \pmb{97.80}$\pm$\pmb{0.05} & 80.37$\pm$0.33 
			& 90.70$\pm$0.65 & 
			90.70$\pm$0.74 &
			91.42$\pm$2.21\\
			Kappa&&89.20$\pm$1.80& 95.44$\pm$0.05 & 85.40$\pm$0.27 & 
			84.66$\pm$1.97 & 91.77$\pm$0.39 &
			\pmb{96.15}$\pm$\pmb{0.73}\\
			\bottomrule 
			
		\end{tabular}
	}
\end{table*}

The classification results we get on the Salinas dataset are listed in Table 
\ref{Table7}. Compared with other algorithms, BLGCN basically achieves the 
highest classification accuracy in all categories, while maintaining high 
stability. We find that the SVM method also has higher classification accuracy 
than the deep learning algorithm, which shows that there is also room for the 
application of traditional classification methods in datasets with sufficient 
and balanced sample data such as Salinas.
\begin{table*}[hptb]\scriptsize 
	\renewcommand\arraystretch{1}
	\centering  
	\caption{\\CLASSIFICATION RESULTS ON SALINAS DATASET}  
	\label{Table7}
	\resizebox{0.8\linewidth}{!}{
		\begin{tabular}{cccccccc} 
			\toprule 
			Class  & & SVM & 3D-CNN(0.3) & GCN & S$^2$GCN & MDGCN & BLGCN\\
			\midrule
			1 & & 98.80$\pm$0.10 & 98.38$\pm$0.04 & 97.44$\pm$0.55 & 
			97.90$\pm$0.43 & 98.26$\pm$0.02
			& \pmb{100.00}$\pm$\pmb{0.00} \\
			2 & & 99.80$\pm$0.11 & 98.88$\pm$0.06 & 99.94$\pm$0.13 & 
			98.07$\pm$0.29 & 98.18$\pm$0.13
			& \pmb{100.00}$\pm$\pmb{0.00} \\
			3 & & 98.94$\pm$0.51 & 99.19$\pm$0.32 & 96.88$\pm$2.42 & 
			96.06$\pm$0.55 & 98.08$\pm$0.15
			& \pmb{100.00}$\pm$\pmb{0.00} \\
			4 & & 99.25$\pm$0.16 & 99.38$\pm$0.17 & 97.42$\pm$0.62 & 
			97.99$\pm$0.51 & 95.81$\pm$0.87
			& \pmb{100.00}$\pm$\pmb{0.00} \\
			5 & & 99.00$\pm$0.00 & 99.11$\pm$0.38 & 89.53$\pm$0.17 & 
			96.46$\pm$0.60 & 96.27$\pm$1.32
			& \pmb{99.11}$\pm$\pmb{1.04} \\
			6 & & 99.91$\pm$0.12 & 99.55$\pm$0.05 &100.00$\pm$0.00 & 
			98.21$\pm$0.19 & 97.40$\pm$1.02
			& \pmb{100.00}$\pm$\pmb{0.00} \\
			7 & & 99.88$\pm$0.15 & 99.50$\pm$0.11 & 99.72$\pm$0.27 & 
			97.95$\pm$0.24 & 96.49$\pm$1.56
			& \pmb{100.00}$\pm$\pmb{0.00} \\
			8 & & 84.77$\pm$0.24 & 77.78$\pm$11.49& 97.68$\pm$0.43 & 
			69.88$\pm$5.89 & 91.18$\pm$2.96
			& \pmb{99.19}$\pm$\pmb{1.04} \\
			9 & & 99.56$\pm$0.10 & 99.75$\pm$0.05 & 99.42$\pm$0.01 & 
			97.22$\pm$0.97 & 98.28$\pm$0.79
			& \pmb{100.00}$\pm$\pmb{0.00} \\
			10& & 96.23$\pm$0.63 & 97.57$\pm$0.05 & 95.84$\pm$0.49 & 
			89.95$\pm$2.41 & 96.62$\pm$1.69
			& \pmb{97.67}$\pm$\pmb{1.93} \\
			11& & 97.81$\pm$1.24 & 97.66$\pm$0.50 & 99.08$\pm$1.94 & 
			96.90$\pm$1.66 & 97.68$\pm$2.04
			& \pmb{99.55}$\pm$\pmb{0.63} \\
			12& & 97.41$\pm$0.61 & 98.69$\pm$0.12 &100.00$\pm$0.00 & 
			98.44$\pm$0.96 & 97.39$\pm$2.57
			& \pmb{100.00}$\pm$\pmb{0.00} \\
			13& & 99.31$\pm$0.60 & 98.61$\pm$0.68 &100.00$\pm$0.00 & 
			96.73$\pm$1.02 & 95.91$\pm$3.19
			& \pmb{100.00}$\pm$\pmb{0.00} \\
			14& & 97.41$\pm$0.64 & 97.86$\pm$0.93 & 95.72$\pm$4.65 & 
			94.67$\pm$1.87 & 96.23$\pm$0.89
			& \pmb{98.97}$\pm$\pmb{0.00} \\
			15& & 71.80$\pm$0.81 & 76.68$\pm$5.87 & 75.94$\pm$9.67 & 
			69.56$\pm$3.58 & 
			\pmb{94.06}$\pm$\pmb{2.09}
			& 91.79$\pm$1.22 \\
			16& & \pmb{99.21}$\pm$\pmb{0.21} & 94.47$\pm$0.18 & 93.69$\pm$3.79 
			& 95.81$\pm$1.03 & 
			95.58$\pm$1.45
			& 94.32$\pm$8.00 \\
			\midrule
			OA& & 92.63$\pm$1.71 & 91.03$\pm$3.10 & 94.75$\pm$1.66 & 
			87.40$\pm$1.24 & 96.52$\pm$2.81
			& \pmb{98.30}$\pm$\pmb{0.22} \\
			AA& & 95.90$\pm$0.03 & 95.88$\pm$1.09 & 96.22$\pm$1.33 & 
			93.14$\pm$0.10 & 96.52$\pm$0.92
			& \pmb{99.02}$\pm$\pmb{0.24} \\
			Kappa&&91.80$\pm$0.20& 90.05$\pm$3.39 & 94.13$\pm$1.86 & 
			86.12$\pm$2.01 & 95.27$\pm$0.96
			& \pmb{98.10}$\pm$\pmb{0.25} \\
			\bottomrule 
			
		\end{tabular}
	}
\end{table*}

We also apply BLGCN on the Houston University dataset to compare the 
classification efficiency. As we can see from Table \ref{Table8}, the algorithm 
proposed in this paper greatly improves the classification accuracy, has 
obvious advantages over other algorithms, and has high classification 
performance on various sample classes. Besides, the comparison algorithms based 
on GCN have the problem that the classification accuracies on the majority 
samples 
classes such as Class 7 $ \sim $ 12 are relatively low due to the minor ratio 
of training set samples. Because of the strong generalization ability in 
Bayesian layer, our model alleviates this problem successfully.
\begin{table*}[hptb]\scriptsize 
	\renewcommand\arraystretch{1}
	\centering  
	\caption{\\ \mbox{CLASSIFICATION RESULTS ON HOUSTON UNIVERSITY DATASET } } 
	\label{Table8}
	\resizebox{0.8\linewidth}{!}{
		\begin{tabular}{cccccccc} 
			\toprule 
			Class  & & SVM & 3D-CNN(0.3) & GCN & S$^2$GCN & MDGCN & BLGCN\\
			\midrule
			1 & & 95.80$\pm$0.60 & 98.88$\pm$0.44 & 81.07$\pm$0.00 & 
			95.41$\pm$2.03 & 92.72$\pm$0.97 &
			\pmb{99.58}$\pm$\pmb{0.70} \\
			2 & & 97.41$\pm$0.51 & 98.84$\pm$0.38 & 
			\pmb{100.00}$\pm$\pmb{0.00}& 97.66$\pm$1.08 & 
			92.97$\pm$0.78 & 99.34$\pm$0.06 \\
			3 & & 99.25$\pm$0.48 & 99.42$\pm$0.44 & 74.89$\pm$2.21 & 
			97.97$\pm$1.35 & 97.39$\pm$0.60 &
			\pmb{100.00}$\pm$\pmb{0.00}\\
			4 & & 96.83$\pm$0.75 & 99.30$\pm$0.35 & 89.24$\pm$0.00 & 
			96.78$\pm$1.14 & 94.87$\pm$1.05 &
			\pmb{100.00}$\pm$\pmb{0.00}\\
			5 & & 95.44$\pm$1.41 & 99.38$\pm$0.29 & 92.34$\pm$3.97 & 
			96.76$\pm$0.97 & 98.27$\pm$0.32 &
			\pmb{99.97}$\pm$\pmb{0.06} \\
			6 & & 96.16$\pm$2.59 & 96.44$\pm$2.26 & 86.90$\pm$0.00 & 
			95.95$\pm$1.45 & 92.58$\pm$1.14 &
			\pmb{98.23}$\pm$\pmb{1.71} \\
			7 & & 83.21$\pm$1.84 & 93.26$\pm$0.89 & 51.54$\pm$1.43 & 
			82.71$\pm$3.54 & 87.01$\pm$2.28 &
			\pmb{97.91}$\pm$\pmb{0.96} \\
			8 & & 72.74$\pm$1.72 & 94.56$\pm$2.89 & 90.73$\pm$0.00 & 
			75.44$\pm$4.11 & 79.83$\pm$4.07 &
			\pmb{94.82}$\pm$\pmb{1.71} \\
			9 & & 76.51$\pm$1.21 & 90.02$\pm$0.76 & 32.06$\pm$7.07 & 
			81.41$\pm$2.69 & 88.96$\pm$0.19 &
			\pmb{92.97}$\pm$\pmb{2.56} \\
			10& & 81.32$\pm$1.40 & 91.90$\pm$0.47 & 65.99$\pm$2.96 & 
			86.05$\pm$2.01 & 89.38$\pm$1.70 &
			\pmb{99.83}$\pm$\pmb{0.07} \\
			11& & 81.74$\pm$1.26 & 92.92$\pm$0.86 & 62.56$\pm$2.42 & 
			87.75$\pm$2.00 & 86.07$\pm$1.64 &
			\pmb{98.62}$\pm$\pmb{0.92} \\
			12& & 69.19$\pm$1.82 & 91.24$\pm$2.09 & 70.63$\pm$3.68 & 
			77.91$\pm$4.68 & 88.76$\pm$2.97 &
			\pmb{99.69}$\pm$\pmb{0.30} \\
			13& & 48.51$\pm$6.95 & 98.94$\pm$4.48 & 0.00$\pm$0.00  & 
			74.93$\pm$4.26 & 92.08$\pm$1.02 &
			\pmb{96.49}$\pm$\pmb{1.29} \\
			14& & 96.25$\pm$1.47 & 98.94$\pm$0.34 & 98.88$\pm$2.51 & 
			98.53$\pm$0.44 & 98.69$\pm$0.39 &
			\pmb{100.00}$\pm$\pmb{0.00}\\
			15& & 98.08$\pm$0.51 & 98.80$\pm$0.48 & 72.44$\pm$4.62 & 
			97.13$\pm$0.68 & 95.55$\pm$0.44 &
			\pmb{100.00}$\pm$\pmb{0.00}\\
			\midrule
			OA& & 85.67$\pm$0.43 & 95.14$\pm$0.51 & 72.93$\pm$0.44 & 
			88.49$\pm$1.57 & 90.71$\pm$1.00 &
			\pmb{98.36}$\pm$\pmb{0.32} \\
			AA& & 86.28$\pm$0.96 & 95.60$\pm$0.71 & 70.44$\pm$0.54 & 
			89.50$\pm$1.04 & 91.68$\pm$1.07 &
			\pmb{98.30}$\pm$\pmb{0.40} \\
			Kappa&&84.71$\pm$0.51& 94.74$\pm$0.53 & 71.73$\pm$0.47 & 
			87.62$\pm$1.46 & 90.02$\pm$0.89 &
			\pmb{98.23}$\pm$\pmb{0.35} \\
			
			\bottomrule 
			
		\end{tabular}
	}
\end{table*}

\subsection{Ablation Study}

We design ablation experiments for multiple functional modules implemented in 
BLGCN, and verify the necessity of each module in order to achieve 
high accuracy in classification. It is also been validated that dynamic control 
effects by processing the Indian Pines dataset.

\subsubsection[1.]{Graph convolution operation}
By using the method of graph convolution, our algorithm effectively extracts 
the spatial adjacency relationship between the superpixels of hyperspectral 
images, which greatly improves the classification efficiency. Among our method, 
the key operation of extracting the spatial relationship is the convolution 
operation of the feature matrix and the adjacency matrix. Therefore, we verify 
the necessity of the graph convolution module by removing the adjacency 
matrix convolution operation. The relevant comparison results are listed in 
Table \ref{Table9}. We can find that after removing the graph convolution 
operation, the overall classification results are greatly affected, and the 
accuracy rates have dropped significantly.

\subsubsection[2.]{Minority class data generation module}
Facing the problem of low accuracy in minority class classification that often 
occurs in practical classification tasks, this algorithm combines the 
generative adversarial network method to achieve effective expansion of 
minority classes. Therefore, we design a comparative experiment of removing the 
minority generation module and directly input the original data into the 
classification module. We list the results of the experiments in Table 
\ref{Table9}, in which the classification accuracies of the minority class 1, 
7, and class 9 have dropped significantly after the minority class data 
generation module is removed, and they cannot be effectively classified at all.

\subsubsection[3.]{Dynamic control training module guided by confidence intervals}
\begin{figure}[t]
	\includegraphics[width=\linewidth]{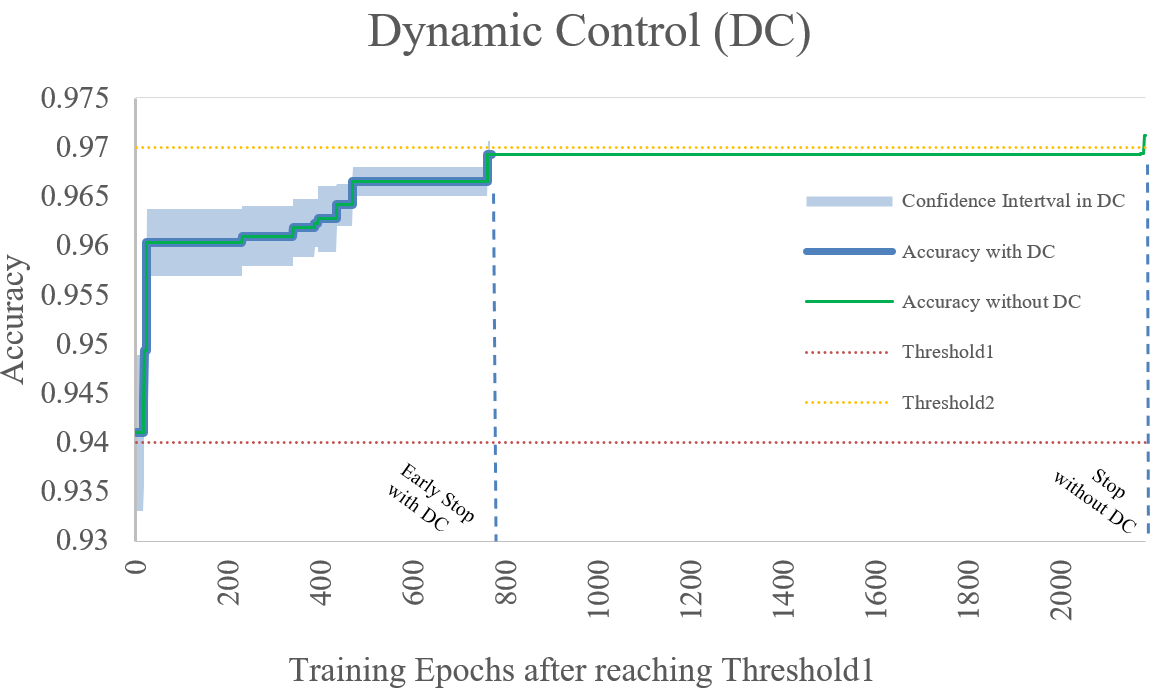}
	\caption{Ablation Study of Dynamic Control} 
	\label{fig_DynamicControl}
\end{figure}

Due to the uncertainty of the Bayesian neural network itself, we combine this 
feature to quantify the output uncertainty by calculating the confidence 
interval of the classification result, and dynamically control the training 
process whether to proceed. We design a comparative experiment on this module 
by removing the uncertainty calculation of the BLGCN classification results and 
the two thresholds in the training process, and we plot the correspondence 
between the accuracy of the validation set and the training batch in Fig. 
\ref{fig_DynamicControl}. It is shown that when given a 95\% confidence level, 
the model automatically stops learning when the confidence interval reaches the 
second threshold, which saves plenty of training time.
\begin{table}[hptb] 
	\renewcommand\arraystretch{1}
	\centering  
	\caption{\\ABLATION STUDY RESULTS ON INDIAN PINES DATASET}  
	\label{Table9}
	\resizebox{3.5in}{!}{
		\begin{tabular}{ccccc} 
			\toprule 
			Class  & & BLGCN without GCN & BLGCN without GAN & BLGCN \\
			\midrule
			1 & & 0.00$\pm$0.00   & 0.00$\pm$0.00   & 
			\pmb{92.30}$\pm$\pmb{3.63} \\
			2 & & 72.82$\pm$18.21 & 94.05$\pm$7.43  & 
			\pmb{96.80}$\pm$\pmb{1.19} \\
			3 & & 59.58$\pm$21.11 & 92.45$\pm$3.12  & 
			\pmb{95.94}$\pm$\pmb{3.27} \\
			4 & & 49.04$\pm$15.69 & \pmb{98.44}$\pm$\pmb{1.00}  & 
			96.92$\pm$6.19 \\
			5 & & 89.76$\pm$2.87  & \pmb{96.85}$\pm$\pmb{1.42}  & 
			93.26$\pm$4.50 \\
			6 & & 94.38$\pm$2.50  & \pmb{99.77}$\pm$\pmb{0.50}  & 
			99.40$\pm$0.80 \\
			7 & & 0.82$\pm$1.73   & 0.00$\pm$0.00   & 
			\pmb{93.30}$\pm$\pmb{7.91} \\
			8 & & 95.90$\pm$3.71  & \pmb{100.00}$\pm$\pmb{0.00} & 
			\pmb{100.00}$\pm$\pmb{0.00}\\
			9 & & 17.50$\pm$28.09 & 0.00$\pm$0.00   & 
			\pmb{100.00}$\pm$\pmb{0.00}\\
			10& & 57.64$\pm$30.88 & 84.42$\pm$4.68  & 
			\pmb{92.19}$\pm$\pmb{4.00} \\
			11& & 69.82$\pm$15.07 & 94.39$\pm$9.73  & 
			\pmb{98.08}$\pm$\pmb{1.21} \\
			12& & 69.54$\pm$5.62  & 94.67$\pm$6.39  & 
			\pmb{97.94}$\pm$\pmb{3.28} \\
			13& & 96.32$\pm$2.86  & \pmb{100.00}$\pm$\pmb{0.00} & 
			99.86$\pm$0.33 \\
			14& & 97.37$\pm$0.50  & \pmb{100.00}$\pm$\pmb{0.00} & 
			99.72$\pm$0.82 \\
			15& & 44.52$\pm$1.60  & 90.81$\pm$2.16  & 
			\pmb{95.07}$\pm$\pmb{3.12} \\
			16& & 90.32$\pm$4.91  & 100.00$\pm$0.00 & 100.00$\pm$0.00\\
			\midrule
			OA& & 74.28$\pm$4.41  & 94.08$\pm$2.19  & 
			\pmb{97.22}$\pm$\pmb{0.81} \\
			AA& & 65.36$\pm$3.17  & 81.85$\pm$0.95  & 
			\pmb{97.02}$\pm$\pmb{1.16} \\
			Kappa&&70.70$\pm$4.74 & 93.23$\pm$2.48  & 
			\pmb{96.81}$\pm$\pmb{0.94} \\
			\bottomrule 
			
		\end{tabular}
	}
\end{table}

\section{Conclusion}
\label{Conclusion}
In this paper, we propose a novel BLGCN method for HSI classification. The 
proposed Bayesian layer is applied to the GCN-based networks to quantify the 
uncertainty of the output results. Since the distribution form is presented on 
only a few parts of the network, the proposed method maintains high 
classification accuracy and generalization ability. 
Meanwhile, newly designed GANs on minority class data are trained to enlarge 
its capacity and solve the sample imbalance problem in HSI dataset. 
To improve the training efficiency, a dynamic control strategy is designed for 
an early termination of the training process when the confidence interval 
reaches the threshold. 
The experimental results on four open source HSI datasets demonstrate the 
superiority of our proposed BLGCN. In addition, ablation studies are arranged 
to verify the contribution of different modules including graph convolution 
operation, data generation module and dynamic control strategy.

In the future, further research will be implemented on the various application 
field of Bayesian layer, including image classification and graph node 
prediction. Moreover, we will continue to improve the theoretical basis of 
Bayesian layer and make it available to various neural networks.

\bibliographystyle{IEEEtran}
\bibliography{bibfile}

\end{document}